\documentclass{article} 
\usepackage{nips13submit_e,times}
\usepackage{graphicx}
\usepackage[colorlinks=true, allcolors=blue]{hyperref}
\usepackage{url}
\usepackage{titlesec}
\usepackage{float}
\titlespacing*{\section}{0pt}{0.1\baselineskip}{0.1\baselineskip}
\titlespacing*{\subsection}{0pt}{0.005\baselineskip}{0.005\baselineskip}
\title{Recurrent and Contextual Models for Visual Question Answering}

\author{
Abhijit Sharang\\
Stanford University\\
\texttt{abhisg@stanford.edu}
\And 
Eric Lau\\
Stanford University\\
\texttt{eclau@stanford.edu}
}

\nipsfinalcopy 

\begin{document}
\maketitle

\begin{abstract}
We propose a series of recurrent and contextual neural network models for multiple choice visual question answering on the Visual7W dataset. Motivated by divergent trends in model complexities in the literature, we explore the balance between model expressiveness and simplicity by studying incrementally more complex architectures. We start with LSTM-encoding of input questions and answers; build on this with context generation by LSTM-encodings of neural image and question representations and attention over images; and evaluate the diversity and predictive power of our models and the ensemble thereof. All models are evaluated against a simple baseline inspired by the current state-of-the-art, consisting of involving simple concatenation of bag-of-words and CNN representations for the text and images, respectively. Generally, we observe marked variation in image-reasoning performance between our models not obvious from their overall performance, as well as evidence of dataset bias. Our standalone models achieve accuracies up to $64.6\%$, while the ensemble of all models achieves the best accuracy of $66.67\%$, within $0.5\%$ of the current state-of-the-art for Visual7W. 

\end{abstract}

\section{Introduction}

Question answering is regarded as a complex task in natural language processing and artificial intelligence in general. A high-performance QA system should demonstrate a wide range of capabilities, such as semantic reasoning, sentiment analysis, and contextual inference of language. Visual question-answering (vQA) is an extension of text-based QA which requires understanding of both images and questions about them. Common methods involve using convolutional and recurrent neural networks to map image/text pairs to some vector space which is representational of the interaction of the image with the text. Several compositional models for combining these multimodal representations have been explored \cite{wu2016visual}. 

In the literature, there exists two divergent trends in state-of-the-art vQA systems. One is toward increasingly complex recurrent models, often with complex attention mechanisms over the images and text. In these models, the textual content often governs which areas of the image are more important for reasoning about the correct answer. 


The other trend is towards much more basic models involving simple features like bag-of-words and models such as the multilayer perceptron (MLP) which achieve performance on current vQA datasets comparable to, and often in  excess of, their more complex counterparts. These studies suggest bias in current visual question answering datasets which allows models to ``guess'' the answer without having to develop a deeper understanding of the question or image context.

In this work, we develop and evaluate recurrent models to systematically combine these image and text features together in incrementally more expressive ways while maintaining model simplicity for better generalization, on the Visual7W dataset. Specifically, we  investigate novel multiple-choice answering variants of a basic  existing  recurrent visual-QA model involving  simple concatenation of LSTM-encoded visual and text features that is fed to a MLP; separate LSTM encodings of question and image features into a context vector to augment the LSTM-encoding of the related answer choice; attention over the question and image; and the overall ensemble of all our models. We evaluate these against a simple baseline in which bag-of-words features for the question and answers are concatenated with the image representation and fed to a MLP, which is inspired by the current state-of-the-art in the literature. In the course of our study, we hope to better characterize the potential dataset biases mentioned previously for Visual7W. 

\section{Background and Related Work}

\subsection{Standard and Visual Question Answering}


Though textual question answering is a well-established standard task in natural language processing, relatively recent improvements in recurrent neural network (RNN) models, as well as convolutional neural network (CNN) models for image recognition, have been effectively applied in combination to vQA \cite{malinowski2015ask}. Several vQA datasets have been released, such as VQA 1.0, Visual Genome, and Visual7W. However, recent studies have shown that some datasets, such as VQA 1.0, exhibit strong language bias \cite{wu2016visual}, and surprisingly few questions require non-trivial reasoning and abstraction to arrive at the answer. Visual7W  \cite{zhu2016cvpr} consists of 327,939 QA pairs and claims to have greater question-answer complexity and diversity to more rigorously evaluate vQA models.  VQA 2.0 is under development  and purports to have improvements in the same vein. 

\subsection{Trend 1: Toward Complex Neural Architectures}

A predictable trend of recent vQA  systems in the literature has been towards more complex neural architectures often involving recurrent models and  attention mechanisms. For example, Shih et al. \cite{shih2016look} attempted to map both the question and answer text and images into a latent space where inner products of textual features yielded attention regions for the associated image.  In addition, Lu et al. \cite{lu2016hierarchical} attempt to create a co-attention model that  performs joint reasoning over the question and image. 

\subsection{Trend 2: Toward Simple Baselines}

Recent work on very simple baselines has resulted in basic systems with performance comparable to or exceeding that of more complex recurrent systems on  VQA 1.0 and Visual7W. For instance, Zhou et al. \cite{zhou2015simple} demonstrated a  simple non-recurrent model where the input question is transformed into a word feature using naive bag-of-words, simply concatenated with deep image features obtained using GoogLeNet, and fed to a softmax layer to predict the answer class. 

Similarly, Jabri et al. \cite{jabri2016revisiting} created a system that instead took the full question-image-answer triplet $(q,a,I)$ as the sample input. In their system, 300-dimensional bag-of-words word2vec features for the question and multiple choice answer are concatenated with 2048-dimensional deep image features from Resnet-101 and are fed to a MLP with 8192 hidden units. Their model achieves an accuracy of $67.1\%$, the current state-of-the-art on Visual7W.  

\section{Approach}

\begin{figure}[h]
\begin{center}
\includegraphics[scale=0.5]{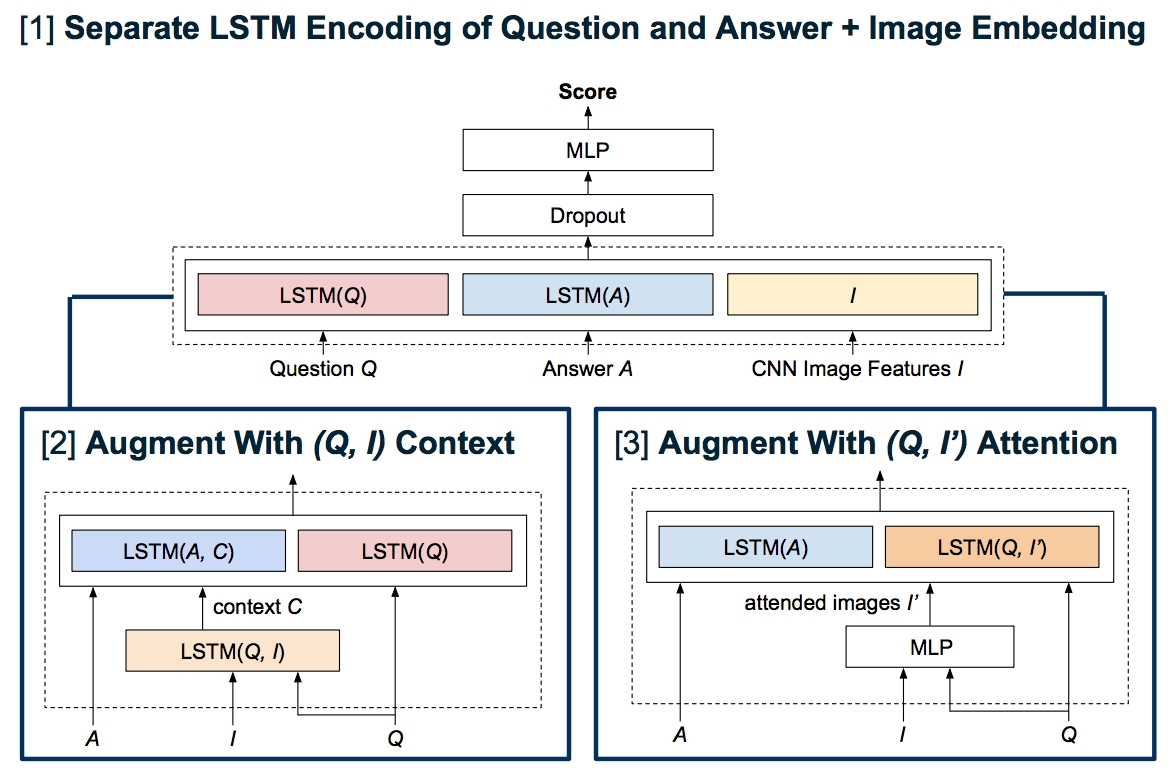}
\end{center}
\caption{Experimental models. In [2] and [3], we replace the dotted portion of [1] with minor adjustments elsewhere. For each LSTM, we use the last hidden state as input to the next layer.}
\label{fig:arch}
\end{figure}

We explore model architectures of incrementally increasing complexity, starting with simple non-recurrent baselines and then incrementally more complex recurrent, contextual, and attentional models which combine some or all of the $(q,a,I)$ triplet for a given question $q$, answer choice $a$ (of which there are four per $q$), and CNN image features $I$, into new intermediate neural feature representations. A diagram of our models is shown in Figure \ref{fig:arch}. For our individual experimental models, we describe relevant modifications leading to the generation of the feature vector only; the rest of the model can be assumed to be the same, minus minor hyperparameter adjustments elsewhere. 

\subsection{Baseline: Simple Non-Recurrent Bag-Of-Words}
The simplest model for the task is an adoption of the bag-of-words (BOW) model in \cite{jabri2016revisiting}. QA text is mapped to word embeddings and images are mapped to representations obtained from a pre-trained CNN after removing the last dense layer. We use {\it Glove-300} for word embeddings and {\it ResNet-50} for image embeddings. Each word representation is 300-dimensional, and each image representation is $7 \times 7 \times 2048$-dimensional. Subsequently, we obtain the BOW representations of the text sequences by averaging the word embeddings over all words in the sequence and the bag-of-images representation by averaging over each $7 \times 7$ stack across $2048$ stacks. These are concatenated together and used as input to a MLP consisting of a fully connected hidden layer and output layer which produces a single score for each $(q,a,I)$ triplet. These scores are then concatenated as a vector of size 4 and normalized using softmax. Categorical cross-entropy is used as the loss function for training.


\subsection{Version [1]: LSTM-Encoding of Question and Image Text}
Version [1] creates recurrent representations of the question and answer sequences. The question and answer word embedding sequences are each passed through their own LSTMs and the last hidden state of each is the vector representation of each sequence. However, we also intend to remain as agnostic to the order of words as possible. For instance ``a cute dog'' and ``a dog which is cute'' should have close mappings in the transformed space. Thus, we use bidirectional LSTMs where the sequences are read forward and  backward and the concatenation serves as the representation.


\subsection{Version [2]: Augmenting With Context From Question and Image}

The baseline and V[1] have minimal interaction between the question image and text and hence may not capture deeper semantic relationships required for proper reasoning. We want a representation where the image embedding affects how the model represents the text and vice-versa. We define the notion of \emph{context}, which is captured by mapping the image and the text to a joint space. 

To obtain the context vector $C$, we transform the question sequence to a sequence \{$(q_w,I)$\}, where $q_w$ is the word embedding representation of a question word $w$ and $I$ is the bag-of-images representation as obtained above. The transformed sequence is passed through a bidirectional LSTM as before, and the final hidden state is used as $C$. We experiment with three architectures which use $C$ to obtain deeper interaction between the text and the image. These models are as follows:
\begin{itemize}
\item The most basic architecture obtains the representation from the bidirectional LSTM on the answer sequence and concatenates it with $C$. This vector is then used as input to the MLP.
\item In the second variant, the input to the MLP consists of the concatenation of the bidirectional LSTM representation of the answer sequence, $C$, and a transformation of $I$. To obtain the image feature transformation, $I$ is fed through a fully-connected dense layer with softmax activation, which reduces its dimension. 
\item In the third variant, we augment the each answer word with $C$ and pass it through its bidirectional LSTM. The augmented answer representation is concatenated with the LSTM-encoding of the question only, as in V[1], and is passed as the feature vector to the MLP. 
\end{itemize}

\subsection{Version [3]: Augmenting With Attention Over Images}

While the context model captures some interaction between the image and the text, it exerts a uniform effect on our models. In reality, we might want the image representation to be influenced by the word representation at each timestep in the sequence. Hence, instead of simply averaging over the $7 \times 7$ values in the 2048-dimensional stack for each word, we want the 2048-dimensional image representation to be dependent on the word with which it is interacting. 

To achieve this, we introduce an attention module. This module transforms the $7 \times 7 \times 2048$-dimensional image embedding to $1 \times 2048$-dimensional embedding in the following manner. Suppose $w$ is the embedding of the word under interaction and $\{I_1,..,I_{2048}\}$ is the raw image embedding. Then, $A_j = MLP([I_j,w])$, where $MLP$ consists of two fully connected dense layers with {\it ReLu} and {\it tanh} activations with the second layer outputting a scalar. Hence, for the text sequence $[t_1,..,t_n]$, we obtain a corresponding image attention sequence $[I'_1,..,I'_n]$, where each $I'$ is a $2048$-dimensional attention representation as calculated above.

We incorporate the attended images by using the sequence of \{$(q_w,I'_w)$\} instead of \{$(q_w,I)$\} as input to one LSTM. We concatenate its last hidden layer with the answer LSTM representation and the transformed image representation as in V[2] as the feature vector to the aforementioned MLP for score generation. 


\section{Experiments}
We use the \emph{Telling} portion of the Visual7W dataset consisting of 69,817 training samples, 28,020 validation samples, and 42,031 test samples. Each sample consists of an image, a question string and four option strings out of which only one choice is correct. To prevent data contamination, each image occurs in exactly one partition of the dataset. The question types are broken down into {\it who, what, when, where, why}, and {\it how}.

In each of the models described above, the architecture becomes the same after the generation of the feature vector. The final two layers consist of a fully connected hidden layer with a large number of hidden units and {\it ReLU} activations and a fully connected output layer with a single unit with sigmoid activations. This copy of the network is run for $(q,a,I)$ triplets over all four answer choices for a given question $q$, and the scores are then normalized by a softmax layer. We experiment with different numbers of units in the hidden layer for each model. To prevent overfitting, we also introduce a dropout layer right after the feature vector and tune the dropout rate; and an $L_2$ regularizer in the output dense layer. Each layer leading to the feature vector in all the models, except the simple BOW model, also has mechanisms for preventing overfitting through dropout and regularization. 

We use categorical cross-entropy to measure the difference between the actual score for a $(q,a,I)$ triplet and the normalized score obtained from the network. Finally, the network is tuned using Adam optimizer \cite{kingma2014adam} for a certain number of iterations with the learning rate, minibatch size, and number of iterations tuned with other hyperparameters. At every step of optimization, the validation accuracy is compared against the best validation accuracy obtained so far, and the snapshot of the model with best accuracy is saved. This model is then used for testing. The entire experimental framework is run with a Tensorflow back-end \cite{tensorflow2015-whitepaper} with front-end code written in Keras \cite{chollet2015keras}. The baseline and all experimental models are our own implementations. 

To handle variable length sequences, we use the \texttt{pad\_sequences} function available in Keras, which prepends the sequences with zeros to make every sequence of equal length. To ensure these zeros do not affect the loss and the gradient, we set \texttt{mask\_zero} flag to True, which ignores any zero in the input during the loss calculation and gradient update.

\subsection{Baseline}
We explore three variants of the BOW baseline. The first version uses the image, the question and the answers. The second version does not use the image, and the third version is only trained on the answers, both of which are useful for studying dataset bias. In all three cases, convergence is reached in 50-60 iterations of training with the default Adam optimizer.

\subsection{Version [1]: LSTM QA-Encoding}
The three variants as described above are used here as well. We tune the hyperparameters for regularization and dropout applied to the matrices $U$ and $W$. Convergence is reached in 90-100 iterations with the default Adam optimizer.

\subsection{Versions [2] and [3]: Context and Attention Augmentations}


Since the more complicated context and attention models have higher susceptibility of overfitting to the training set, the dropout and regularization parameters used in these models is more severe. Also, these models have a more unpredictable optimization landscape and thus the Adam optimizer used for these models has a learning rate of $10^{-4}$, which is 0.1 times the learning rate of the default optimizer. 
We generally observe these models take a much longer to converge and have lower validation accuracy at convergence than the simple models. 


\subsection{Ensemble Model}
Ensembling is an effective technique to obtain a stronger model from a series of weaker models. It works best when models exhibit diversity, that is, different models focus on different aspects of the data and together produce an ensemble model which can produce robust results on each aspect. We hypothesize that our selection of models with increasing complexity is diverse enough to produce a strong ensemble model. For this purpose, we pass each test example through each of the ten models and choose the option chosen by the majority of models as the answer to the question, while breaking ties arbitrarily. We do not perform any training on the ensemble and do not weigh the models. Hence, each model's vote is counted equally. 



\section{Results} 

We exploit our step-by-step progression in additional model complexity to explore their diversity in terms of general and specific performance enhancements and limitations. Results for the array of models is shown in Table \ref{tab:a}. A comparison of the ensemble of our models with other models in the literature for the Visual7W dataset is shown in Table \ref{tab:b}. 

\begin{table}
\begin{center}
\begin{tabular}{l | c | c | c | c | c | c | c}
\hline
Model &\textbf{What}& \textbf{Who}& \textbf{When} &\textbf{How} & \textbf{Where}&\textbf{Why} & \textbf{Overall}\\ \hline
\textbf{Baseline models} & & & & & & &\\
BOW(A) &0.493&0.64&0.763&0.52&0.59&0.583&0.546\\ 
BOW(Q,A) &0.565&0.668&0.78&0.533&0.621&0.621&0.593\\
BOW(Q,A,I) &0.607&0.704&\textbf{0.817}&0.528&0.717&0.63&0.634 \\
\hline
\textbf{Simple models}& & & & & & &\\
BiLSTM(A) &0.494&0.647&0.76&0.514&0.59&0.566&0.545 \\
BiLSTMs(Q,A) &0.566&0.674&0.77&0.553&0.621&0.614&0.596 \\
BiLSTMs(Q,A) + I &0.623&0.701&0.815&0.562&0.721&0.621&0.646 \\ \hline
\textbf{Contextual Models}& & & & & & &\\
LSTMs(A)-Context&0.588&0.685&0.788&0.557&0.677&0.625&0.619 \\
LSTMs(A)-Context + I&0.596&0.701&0.812&0.532&0.72&0.616&0.628\\
LSTMs(Q,A,I)-Context&0.61&0.695&0.815&0.562&0.714&0.627&0.639 \\
LSTMs(Q,A,I')-Attention&0.51&0.597&0.742&0.522&0.61&0.528&0.548\\ \hline
Ensemble-All &\textbf{0.641}&\textbf{0.721}&0.814&\textbf{0.589}&\textbf{0.734}&\textbf{0.67}&\textbf{0.667} \\
\hline
\end{tabular}
\caption{Test accuracies across all models, broken down by question type.}
\label{tab:a}
\end{center}
\end{table}

\begin{table}
\begin{center}
\begin{tabular}{l | c | c | c | c | c | c | c}
\hline
Model &\textbf{What}& \textbf{Who}& \textbf{When} &\textbf{How} & \textbf{Where}&\textbf{Why} & \textbf{Overall}\\ \hline
Fritz et al.\cite{malinowski2015ask} & 0.489 & 0.581 & 0.713 & 0.503 & 0.544 & 0.513 & 0.521\\ 
Zhu et al. \cite{zhu2016cvpr} &  0.515 & 0.595 & 0.75 & 0.498 & 0.57 & 0.555 & 0.556\\
Jabri et al. \cite{jabri2016revisiting} & \textbf{0.645} & \textbf{0.729} & \textbf{0.821} & 0.564 & \textbf{0.759} & \textbf{0.68} & \textbf{0.671} \\
Ensemble model & 0.641&0.721&0.814&\textbf{0.589}&0.734&0.67&0.667\\
\hline
\end{tabular}
\caption{Comparison of the ensemble model with other models in the literature on Visual7W. The other results are from \cite{jabri2016revisiting}.}
\label{tab:b}
\end{center}
\end{table}

\subsection{Effect of Additional Complexity: Context and Attention}

\begin{figure}
\centering
  \includegraphics[scale=0.3]{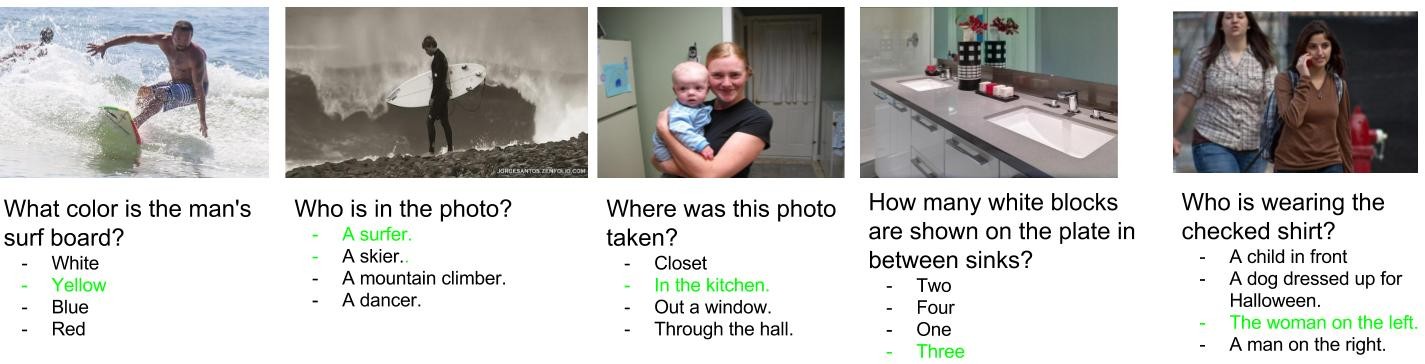}
  \caption{Examples of questions which only the attention model answers correctly.}
  \label{fig:a}
\end{figure}

\begin{figure}
\centering
  \includegraphics[scale=0.35]{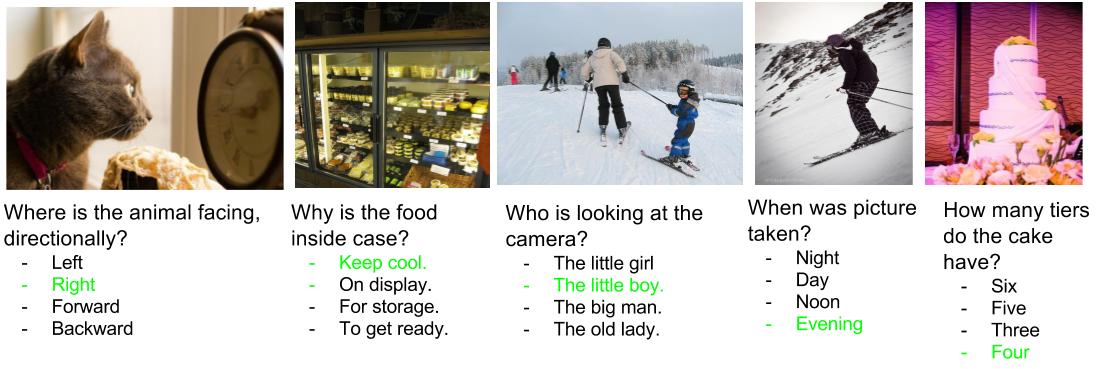}
  \caption{Examples of questions which only the context models answer correctly.}
  \label{fig:b}
\end{figure}

We observe evidence of greater image reasoning for our models augmented with context and attention. To obtain further insight, we explore questions that only these models classify accurately. From the questions for which these models are the ``experts'', we note that most seem to focus on aspects of the image relating to color and spatial positioning. Figures \ref{fig:a} and \ref{fig:b} illustrate some questions which only the model with attention and the models with context answer correctly, respectively. Answering these questions requires focusing on specific sections of the image and understanding the underlying semantics, which is the goal of these enhanced models. For example, the question "How many white blocks are shown on the plate in between sinks?" requires reasoning about the position of the sink, recognizing the blocks, and subsequently filtering out only the white blocks. The attention model is able to capture this relationship. Similarly, the question "When was picture taken?" requires context generation by associating fading light with evening, which the contextual models are able to do. 

Interestingly, we find that simply augmenting the answer and question LSTM-encoded feature inputs to the MLP for the contextual and attentional models, respectively, and keeping the default image embeddings from V[1] is less effective than removing the image embedding and using the augmentations alone with the original LSTM-encodings of the question and answer, respectively. This points to useful subtle interactions between each pair of inputs that can be easily ``masked'' by the concatenation of a high-dimensional image feature representation. 

While we observe increased image reasoning for V[3], it performs worse overall across all question types. However, the overall accuracy is commensurate with that reported by Zhu et. al. \cite{zhu2016cvpr} in their more complicated implementation of attention. Generally, however, the best individual model was V[1], which is evidence of trend (2) seen in the literature towards simpler architectures with high performance, alluding to strong dataset bias as we explore subsequently.


\subsection{Effect of Ensembling}

\begin{figure}
\centering
  \includegraphics[scale=0.35]{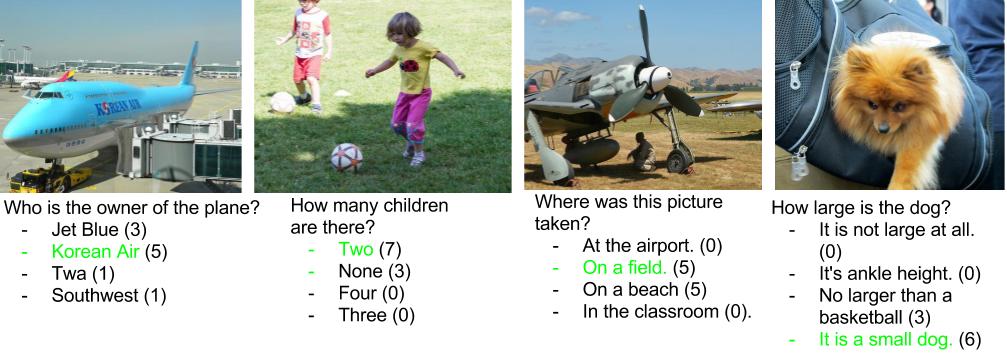}
  \caption{Examples of questions that the ensemble answers correctly. The parenthesized number for each option indicates the number of constituent models voting for it. }
  \label{fig:c}
\end{figure}

Our qualitative and quantitative observations about the diversity of our models are supported by the improved performance of our ensemble, which successfully exploits the diversity across its constituent classifiers. In each question category, except the ``when'' category, there is a marked improvement in the accuracy of around 2-3\% in the ensemble when compared with the accuracy of the best singleton model, indicating that different models specialize at different aspects of the data and their combination has strength across all aspects. The ``when'' category does not show a marked improvement because of higher homogeneity of questions in the category, which is also demonstrated by our subsequent dataset bias analysis. Examples of questions answered correctly by the ensemble are shown in Figure \ref{fig:c}. 
\subsection{Evidence of Dataset Bias}

\begin{figure}[!htb]
\begin{center}
\includegraphics[scale=0.2]{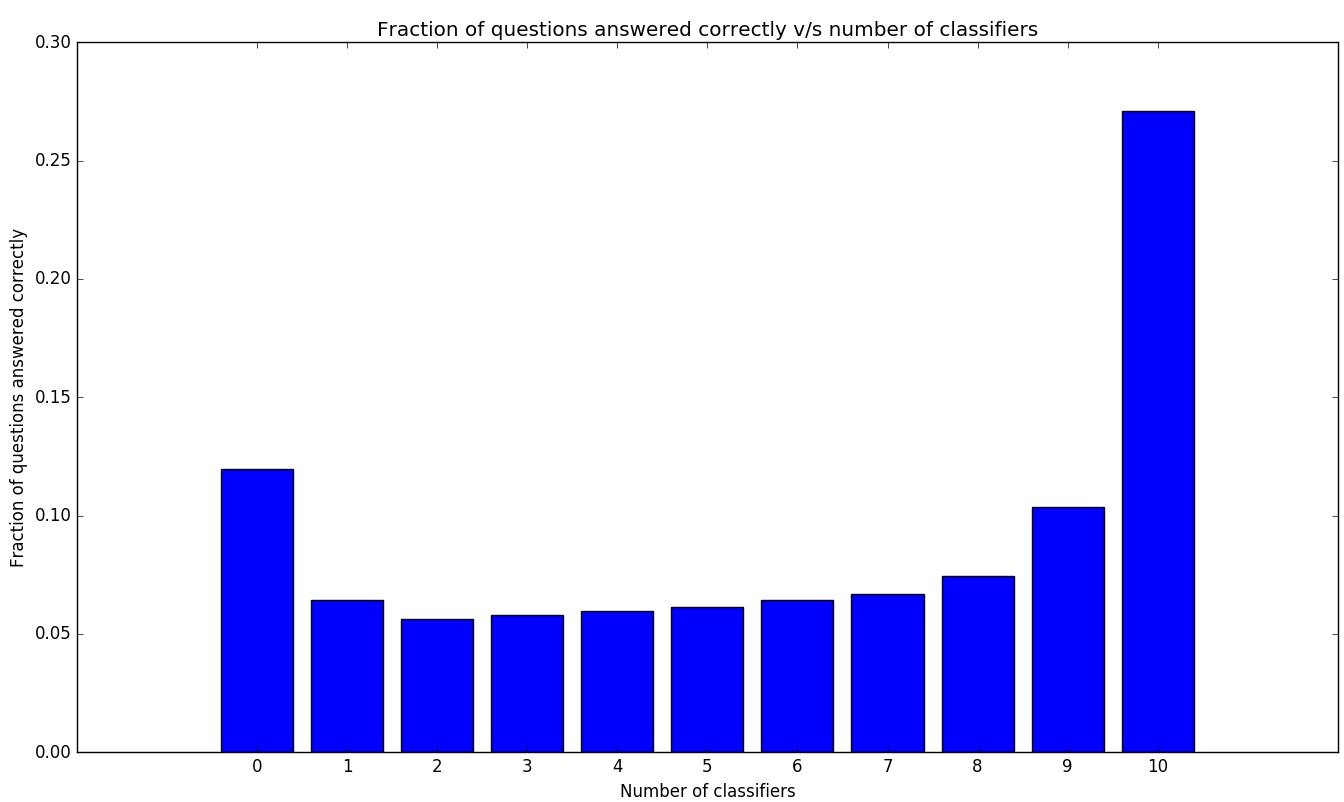}
\end{center}
\caption{Visual7W dataset biases across all question types. Around 28\% of questions are answered correctly by all models while around 13\% of questions are not correctly answered by any classifier. }
\label{fig:d}
\end{figure}

We see that in Figure \ref{fig:d}, there are substantial percentages of questions that all or none of the classifiers answer correctly. Predictably, we see that these ``easy'' questions can be answered without knowledge of the image at all, while ``hard'' questions that all classifiers fail to answer correctly require more in-depth reasoning about the image. Table \ref{tab:d} shows a breakdown of hard questions (answered correctly by less than 3 models), easy questions (answered correctly by more than 7 models) and fair questions (answered correctly by 3-7 models). The ``who'', ``when'', ``where'' and  ``why'' categories have substantial number of easy questions, with the ``when'' category having a percentage as high as 71.3\%. This points at homogeneity in the questions and bias in the text which can picked up by models which do not include images. On the other hand, 33.5\% and 25.1\% of questions in the ``how'' and ``what'' categories, respectively, are answered correctly by very few models.

\begin{figure}[!htb]
\begin{center}
\includegraphics[scale=0.2]{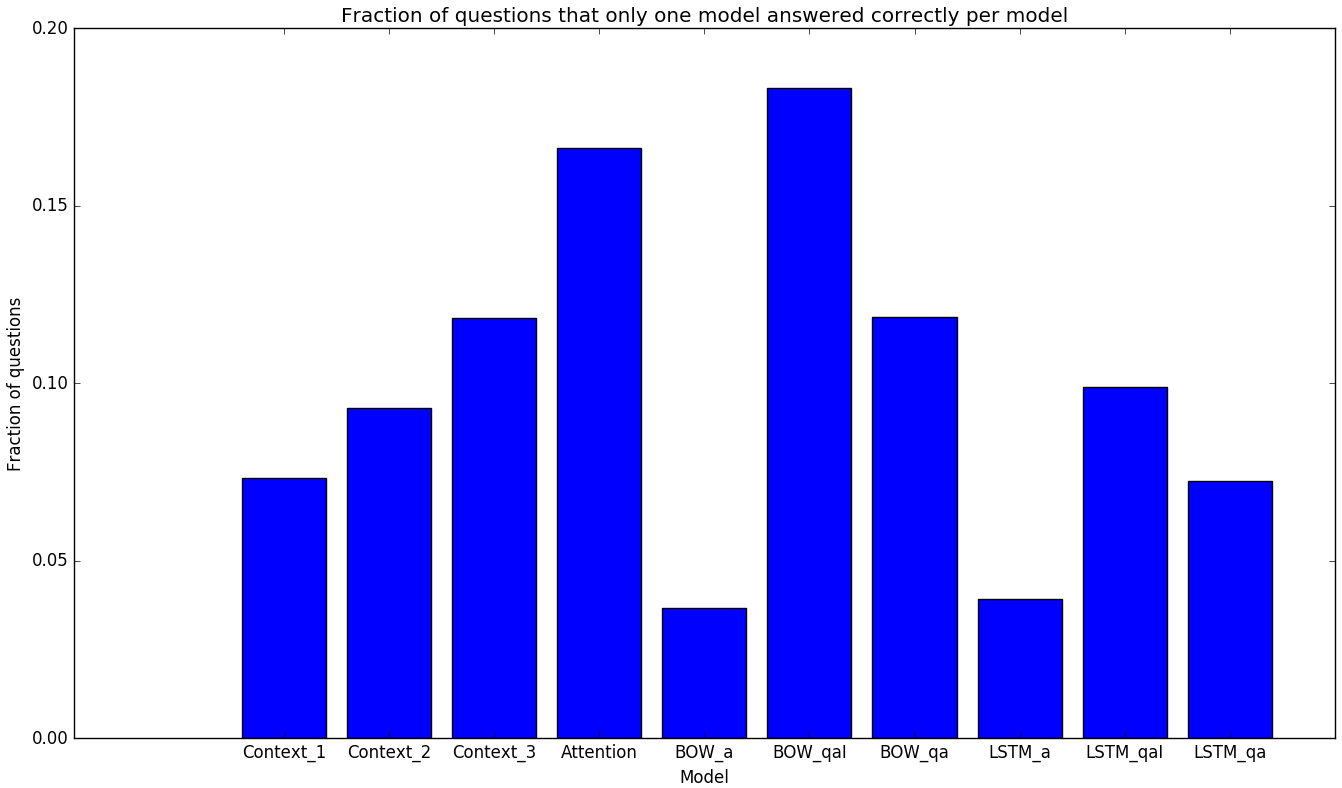}
\end{center}
\caption{Fraction of ``hard'' questions that were answered correctly only by one model, across all models. The attention and simple baseline answer 16.5\% and 18\% of such questions, respectively.}
\label{fig:e}
\end{figure}

Figure \ref{fig:e} shows the distribution of questions that were answered correctly by only one model. The attention and simple baseline have the highest fractions of such questions (16.5\% and 18.5\%) respectively, which demonstrates that the very complex and very simple models perform well on certain tasks.
Hence, while sophisticated image-reasoning models are bound to perform better on ``hard" questions if trained properly, the effect of bias is strong enough for simpler models to dominate the overall performance by maximizing correct answers on ``easy" questions. Future versions of this dataset could factor these bias observations into consideration.


\begin{table}[!htb]
\begin{center}
\begin{tabular}{|l | c | c | c |}
\hline
Category &\textbf{Hard questions}& \textbf{Fair questions}& \textbf{Easy questions} \\\hline
What &0.251 &0.359&0.389\\
Who &0.198&0.241&0.563\\
When&0.132&0.157&0.713\\
How&0.335&0.248&0.417\\
Where&0.18&0.31&0.51\\
Why&0.238&0.305&0.458\\\hline
\end{tabular}
\caption{Fraction of easy, fair and hard questions for each question type.}
\label{tab:d}
\end{center}
\end{table}





\section{Conclusion}

In summary, we have proposed a series of recurrent, contextual, and attentional models for visual question answering that explore the effect of the balance between model expressiveness and simplicity on performance. While we observe that our simplest experimental model involving simple LSTM-encodings of the QA text achieves the best individual model performance, confirming a trend towards simpler architectures found in the literature, we also see that our more complex contextual and attentional models demonstrate noticeable improvements in image reasoning. This is exploited in the improved performance of our ensemble model, which achieves a test accuracy within $0.5 \%$ of state-of-the-art on Visual7W. In evaluating the performance of our series of models, we find evidence of dataset bias in Visual7W, in the form of significant percentages of both universally easy and hard questions that all or none of our models correctly answer, respectively. Future studies would involve evaluating our trained models on other current visual question answering datasets such as VQA 1.0 and 2.0; and training on VQA and evaluating on V7W to explore inter-dataset biases and whether our models generalize; and also investigating more complex models of image attention. 

\subsubsection*{Acknowledgments}

The authors would like to thank  Danqi Chen for her advice and help for this project; and Dr. Manning and Dr. Socher for their instruction. 

\bibliographystyle{unsrt}
\bibliography{nips2013}

\subsubsection*{Appendix}

\begin{figure}[H]
\begin{center}
\includegraphics[scale=0.3]{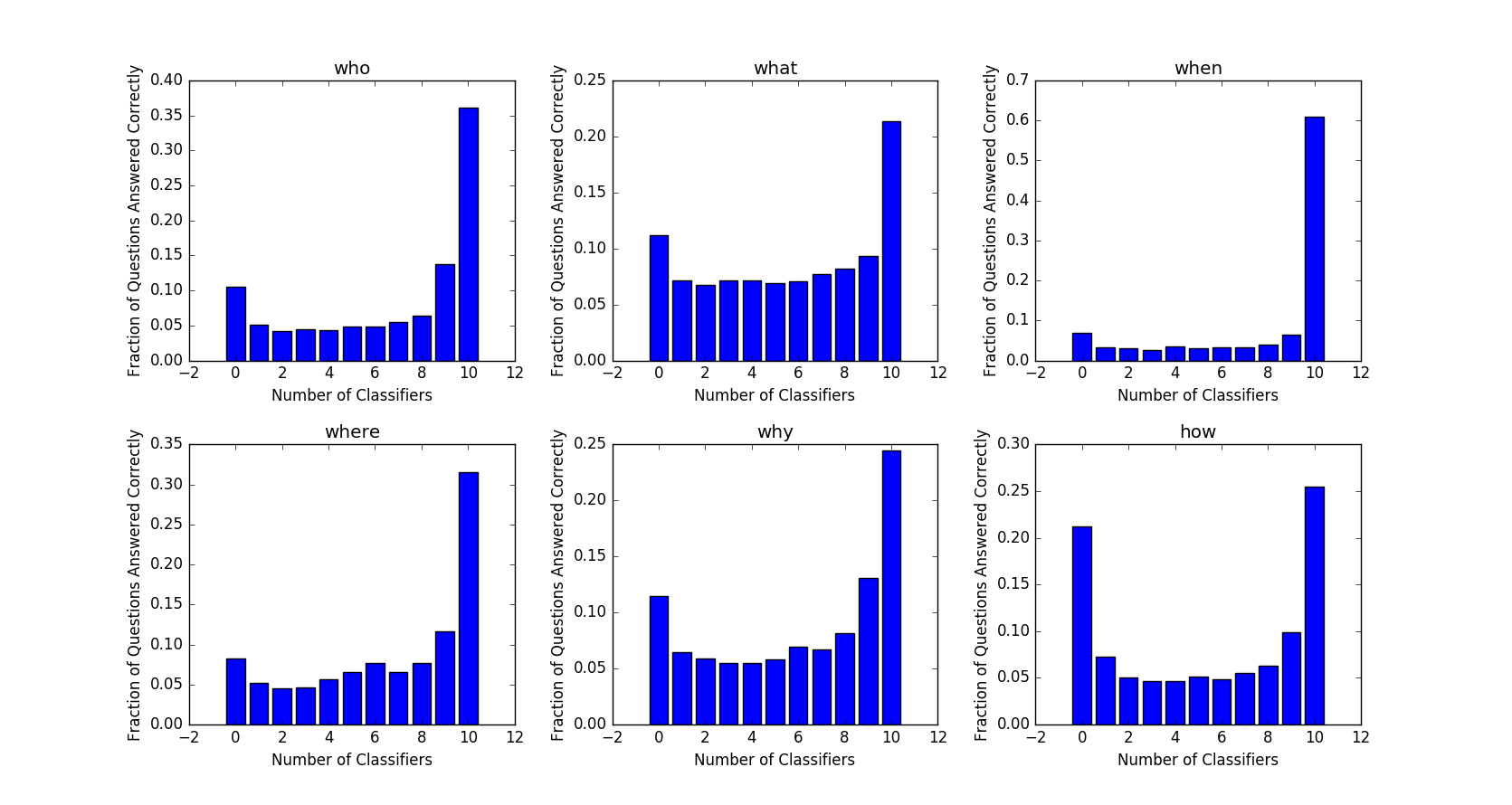}
\end{center}
\caption{Visual7W dataset biases broken down by question type.}
\end{figure}

\end{document}